\documentclass[runningheads]{llncs}
\usepackage{graphicx}
\usepackage[T5]{fontenc}
\usepackage[utf8]{inputenc}
\usepackage{cite}
\usepackage{amssymb}
\usepackage{amsmath}
\usepackage{multirow}
\usepackage{url}
\usepackage{tabularx}
\usepackage{pgfplots}
\usepackage{float}
\usepackage{sidecap}
\usepackage{wrapfig}
\usepackage{hyperref}
\usepackage{makecell}
\usepackage{appendix}
\usepackage{longtable}
\usepackage{caption}
\usepackage{adjustbox}

\begin{document}

\title{Conversational Machine Reading Comprehension for Vietnamese Healthcare Texts}

\author{
Son T. Luu\inst{1,2,\clubsuit} \and
Mao Nguyen Bui \inst{1,2,+} \and
Loi Duc Nguyen \inst{1,2,+} \and
Khiem Vinh Tran \inst{1,2,+} \and
Kiet Van Nguyen\inst{1,2,\clubsuit}\thanks{Corresponding author: Kiet Van Nguyen. Email: kietnv@uit.edu.vn, ORCID: \url{https://orcid.org/0000-0002-8456-2742}} \and
Ngan Luu-Thuy Nguyen\inst{1,2,\clubsuit}}

\institute{University of Information Technology, Ho Chi Minh City, Vietnam \and
Vietnam National University Ho Chi Minh City, Vietnam \\
\email{\inst{\clubsuit}\{sonlt,kietnv,ngannlt\}@uit.edu.vn, \inst{+}\{16520724, 16521722, 17520634\}@gm.uit.edu.vn}}

\newcommand{\footref}[1]{%
    $^{\ref{#1}}$%
}
\newcolumntype{L}[1]{>{\raggedright\let\newline\\\arraybackslash\hspace{0pt}}m{#1}}
\newcolumntype{C}[1]{>{\centering\let\newline\\\arraybackslash\hspace{0pt}}m{#1}}
\newcolumntype{R}[1]{>{\raggedleft\let\newline\\\arraybackslash\hspace{0pt}}m{#1}}

\definecolor{lightblue}{HTML}{4f81c5}

\titlerunning{ }
\authorrunning{ } 
\maketitle              

\begin{abstract}
Machine reading comprehension (MRC) is a sub-field in natural language processing that aims to assist computers understand unstructured texts and then answer questions related to them. In practice, the conversation is an essential way to communicate and transfer information. To help machines understand conversation texts, we present UIT-ViCoQA, a new corpus for conversational machine reading comprehension in the Vietnamese language. This corpus consists of 10,000 questions with answers over 2,000 conversations about health news articles. Then, we evaluate several baseline approaches for conversational machine comprehension on the UIT-ViCoQA corpus. The best model obtains an F1 score of 45.27\%, which is 30.91 points behind human performance (76.18\%), indicating that there is ample room for improvement. Our dataset is available at our website: \url{http://nlp.uit.edu.vn/datasets/} for research purposes.



\end{abstract}

\section{Introduction}
\label{introduction}
Conversation is a standard method to communicate between people, and it plays an important role in human daily life. The process of asking a question and responding to an answer brings helpful information about a specific domain. 

Healthcare is one of the most concerning problems for many people. Many audiences often read the healthcare news, and people tend to discuss frequently about health and medicine. Thus, based on the conversations about healthcare, we constructed a corpus named UIT-ViCoQA for conversational question answering on healthcare texts in Vietnamese. The UIT-ViCoQA contains 2,000 conversations and 10,000 questions from articles about health news in Vietnamese. This corpus is used to train the computer for understanding the conversation and giving the right answers based on the conversation context from questions of users. Besides, we implement neural-based models for conversational question answering including: DrQA \cite{chen-etal-2017-reading}, GraphFlow \cite{ijcai2020-171}, FlowQA \cite{huang2019flowqa}, and SDNet \cite{zhu2019sdnet} on the UIT-ViCoQA corpus. Then, we evaluate the performance of those models on the UIT-ViCoQA dataset.

The main contribution in this paper includes providing a corpus for conversational machine comprehension about healthcare texts in Vietnamese and evaluating the performance of baseline MRC models on the dataset. Our paper is structured as described. Section \ref{related_works} takes a literature review about the conversation machine comprehension corpora and models. Section \ref{corpus} provides overview information about the UIT-ViCoQA dataset. Section \ref{method} introduces available state-of-the-art approaches for the conversational machine comprehension task. Section \ref{experiment} shows our empirical results and error analysis of question-answering models on the UIT-ViCoQA corpus. Finally, Section \ref{conclusion} concludes our works.
\section{Related Works}
\label{related_works}
Machine reading comprehension (MRC) is a challenging task of natural language processing (NLP) which enables machines to understand the reading text and answer the questions \cite{rajpurkar-etal-2016-squad}. Many of MRC corpora are constructed on specific domains, and open domains in English such as SQuAD \cite{rajpurkar-etal-2016-squad} (extractive MRC) on Wikipedia articles, RACE \cite{lai-etal-2017-race} (multiple choices MRC) on High school students English Exams domain, and NarrativeQA \cite{kocisky-etal-2018-narrativeqa} (abstractive MRC) on books and stories domain. For the Vietnamese language, the UIT-ViQuAD \cite{nguyen-etal-2020-vietnamese} (Wikipedia domain), and UIT-ViNewsQA \cite{vannguyen2020new} (Health news domain) are two extractive MRC corpora for machine reading comprehension. Besides, the ViMMRC \cite{9247161} is the multiple-choice reading comprehension corpus on the Vietnamese students' textbook for primary schools domain. 

Machine reading comprehension applied in question-answering (QA) systems is another challenge that the MRC models have to understand both given texts and conversational context and then answer relevant questions. These questions are often paraphrased, contain co-reference queries, and their answers can be spans texts or free-form. This type of MRC is called Conversational Machine Comprehension (CMC) \cite{gupta-etal-2020-conversational}. CoQA \cite{reddy-etal-2019-coqa} and QuAC \cite{choi-etal-2018-quac} are two CMC corpora in English. Based on the CoQA works, we constructed the UIT-ViCoQA for automated reading comprehension on the health news articles in the Vietnamese language. 

Attention-based reasoning with sequence models and FLOW mechanism are two approaches for CMC models, according to Gupta et al. \cite{gupta-etal-2020-conversational}. DrQA\cite{chen-etal-2017-reading} and PGNet\cite{see-etal-2017-get} are two neural attention-based models implemented in the CoQA corpus. Next, SDNet\cite{zhu2019sdnet} is another attention-based model that combines inter-attention and self-attention to comprehend the conversation context. Finally, FlowQA \cite{huang2019flowqa} and GraphFlow \cite{ijcai2020-171} are two flow-based models that used to yield the contextual information through sequences. 
\section{The Corpus}
\label{corpus}
\begin{figure}[H]
    \centering
    \resizebox{0.8\textwidth}{!}{
    \includegraphics[height=3cm,width=3cm]{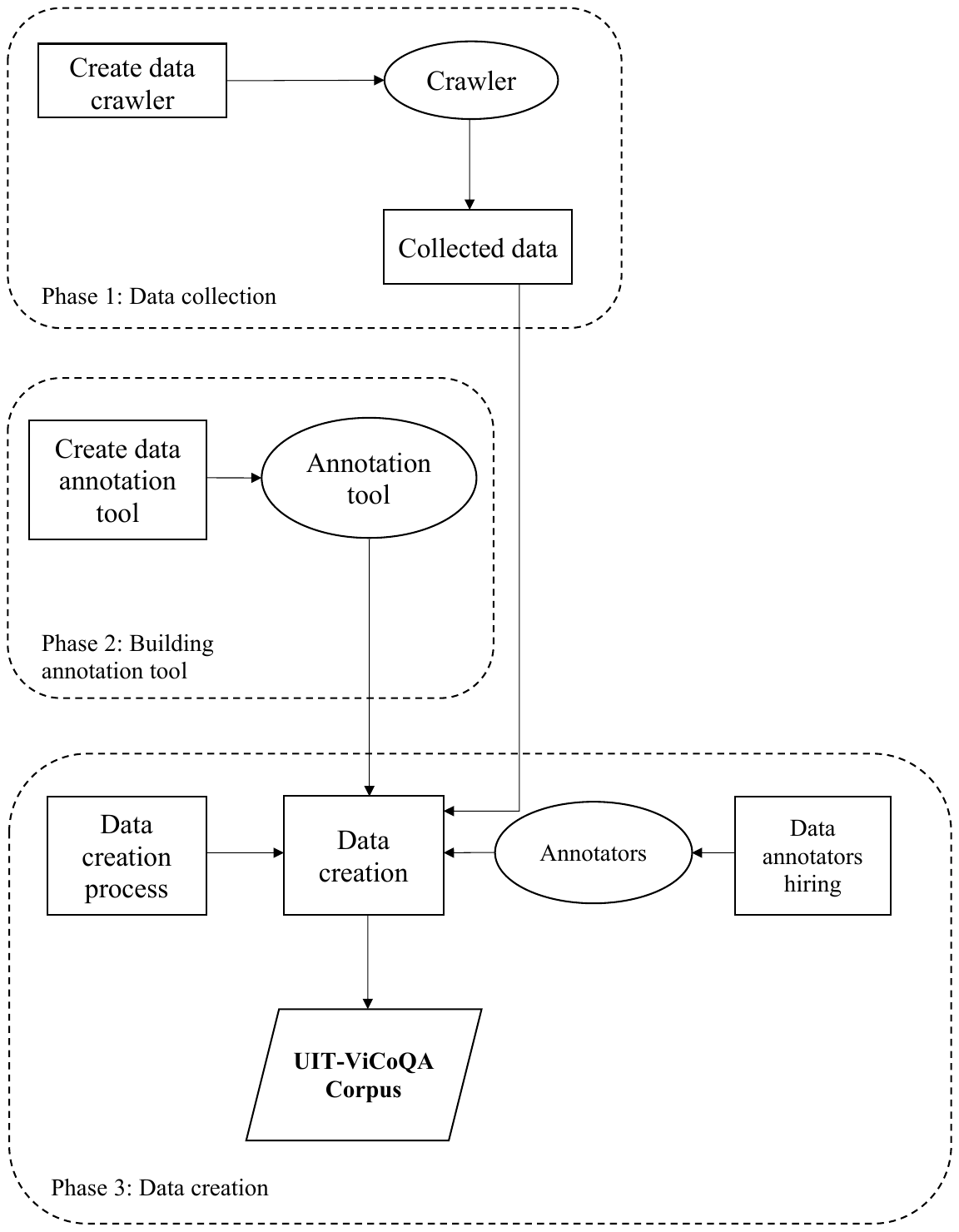}
    }
    \caption{The creation process of the UIT-ViCoQA corpus.}
  \label{fig_data_creation}
\end{figure}

Our data creation process consisting of three phases is described in Figure \ref{fig_data_creation}. In the first phase, we collect news articles about health from VnExpress\footnote{\url{https://vnexpress.net/suc-khoe}} - the most read online newspapers in Vietnam by using scrapy\footnote{\url{https://scrapy.org/}} - a web crawler tool for collecting articles from the online newspaper. In the next phase, we construct an annotation tool for creating conversational data. Our annotation tool allows two annotators to create the conversation based on the given articles. Finally, in the third phase, we hire a team of annotators who create data on our annotation tool. The detailed steps from the annotation process are described below. 

\subsection{Data collection}

For each conversation (C), we hire two different annotators, which are questioners and answerers, respectively. The questioner goes first by asking a question (Q). The question is sent to the answerer then. After receiving the question, the answerer gives the answer by selecting a span of text from the article (S) and then submits the natural answer (A). Next, the annotation system compares the answer given by the answerer with the asked question of the questioner by character level. If the given answer matches about 70\% with the asked question, it is a valid answer, and two annotators can move to the next turn. In contrast, the answerer must give another answer. There is a total of five turns for asking and answer per article. 

In the data creation process, we have some requirements for questioners and answerers as: (1) The answers must be extracted from the article. Questions that cannot be answered according to the article are not allowed, (2) Questioners are encouraged to give questions with synonyms, opposite words, and coreference, and (3) The answers should be short and limited to use new words from the article content. Moreover, the selected answerers need to give full answers with complete texts, correct syntax, and punctuations. 

\subsection{Dataset overview}
\begin{table}[H]
    \centering
    \caption{An example of conversation in the UIT-ViCoQA corpus}
    \label{corpus_example} 
    \resizebox{\textwidth}{!}{
        \begin{tabular}{|l|p{14.2cm}|}
        \hline
        \multicolumn{2}{|p{14.7cm}|}{Trạng thái "ngủ" là cách các tế bào ngay lập tức thay đổi để kháng lại phương pháp điều trị. Các phương pháp điều trị ung thư vú thường thành công, tuy nhiên một số trường hợp ung thư tái phát và tiên lượng xấu hơn.
        Ông Luca Magnani, Khoa Dược, Đại học Hoàng Gia London, Anh, cho biết phương pháp điều trị bằng hormone hiện được sử dụng cho phần lớn bệnh nhân ung thư vú ... 
        (The status of "sleep" is the way when the cell changes immediately to resist treatment. The treatment methods of breast cancer are often successful. However, some cases of cancer recur, and the prognosis worsens.
        Mr. Luca Magnani, Faculty of Medicine, Imperial College London, says that the treatment method by using hormones is used for a huge amount of breast cancer patients ... )
        } \\
        \hline
        Q1 & Phương pháp thường được sử dụng để chữa trị ung thư vú là gì ? (What is the treatment method usually use for breast cancer treatment?) \\
        S1 & Ông Luca Magnani, Khoa Dược, Đại học Hoàng Gia London, Anh, cho biết phương pháp điều trị bằng hormone hiện được sử dụng cho phần lớn bệnh nhân ung thư vú . (Mr. Luca Magnani, Faculty of Medicine, Imperial College London, says that treatment method by using hormone is used for a huge amount of breast cancer patients.) \\
        A1 & điều trị bằng hormone (using hormone)  \\
        \hline
        Q2 & Các bác sĩ có lo ngại gì về phương pháp này? (What are doctors concerned about for this treatment?) \\
        S2 & Từ lâu, các nhà khoa học đã đặt câu hỏi, liệu pháp này thực chất có tiêu diệt được các tế bào ung thư vú không, hay chỉ là chuyển các tế bào sang trạng thái "ngủ yên". (Scientists have long questioned whether this therapy actually kills breast cancer cells, or just puts the cells in an "inactive" state.) \\
        A2 & nó đưa các tế bào ung thư sang trạng thái "ngủ yên" (This treatment puts the cells in an "inactive" state)\\
        \hline
        Q3 & Vậy những nghiên cứu này có ý nghĩa như thế nào? (What profits from these studies?)\\
        S3 & cũng giải thích rằng những phát hiện hiện tại sẽ mở ra lộ trình mới cho việc nghiên cứu chữa trị ung thư. (explaining that current works can open new future researchs about cancer treatments)\\
        A3 & mở ra lộ trình mới cho việc nghiên cứu chữa trị ung thư (Opening new research for cancer treatments) \\
        \hline
        \end{tabular}
    }
\end{table}

The UIT-ViCoQA corpus contains 2,000 conversations. Each conversation consists of a reading article and five question-answer pairs. We follow the structure of the CoQA \cite{reddy-etal-2019-coqa} for our dataset. According to Table \ref{corpus_example}, to answer question Q2, the answerer needs to read the passage and looks back to question Q1 and answer A1 to retrieve the relevant information. Similar to question Q2, the answerer needs to read the reading passage and two previous question-answer pairs (Q1, A1) and (Q2, A2) to extract the answer A3. The chain of question-answer pairs Q1-A1, Q2-A2 is the history of the conversation.

Table \ref{corpus_overview} provides the overview of the UIT-ViCoQA corpus and compares it with the CoQA corpus. The result illustrates that although the number of questions and answers in the UIT-ViCoQA corpus is lower than the CoQA corpus, the average number of words in the UIT-ViCoQA dataset is larger than the CoQA dataset. This is because the interrogative words in English contain a single word (e.g., who?, when?, and why?) while they may have two words in Vietnamese. For example, the words "who" means "ai", "when" means "khi nào" and "why" means "tại sao". Besides, the UIT-ViCoQA is constructed on a specific domain. Hence it is not as diverse as the CoQA corpus.

\begin{table}[H]
    \centering
    \caption{Overview information about the UIT-ViCoQA and CoQA corpus}
    \label{corpus_overview}
    \begin{tabular}{|L{5cm}|R{3cm}|R{3cm}|}
    \hline
    & \textbf{UIT-ViCoQA} & \textbf{CoQA} \\
    \hline
    Domain text & Health domain & Diverse domains \\ 
    \hline
    Number of passages & 2,000 & 8,399  \\
    \hline
    Number of questions & 10,000 & 127,000 \\
    \hline
    Passage length  & 404.1 & 271.0  \\
    \hline
    Question length & 9.4   & 5.5 \\ 
    \hline
    Answer length   & 9.7   & 2.7  \\ 
    \hline
    \end{tabular}
\end{table}

\subsection{Dataset analysis}

\begin{table}[H]
    \caption{The types of question in the UIT-ViCoQA corpus}
    \label{question analysis}
    \resizebox{\textwidth}{!}{
    \begin{tabular}{|L{1.6cm}|L{12.3cm}|R{1cm}|}
        \hline
        \textbf{Question types} & \textbf{Example} & \textbf{Ratio (\%)} \\ \hline
        \makecell[l]{What}  & \makecell*[{{L{12.2cm}}}]{trans fat \textbf{là gì}?  (what is trans fat?)} & 32.6 \\ 
        \hline
        \makecell[l]{How many} & \makecell*[{{L{12.2cm}}}]{Vietnam có \textbf{bao nhiêu} ca nhiễm COVID-19? (\textbf{How many} cases of COVID 19 are detected in Vietnam?) }  & 17.2 \\ \hline
        \makecell[l]{How} & \makecell*[{{L{12.2cm}}}]{DCVax hoạt động như \textbf{thế nào}? (How does DCVax work?)} & 7.6 \\ 
        \hline
        \makecell[l]{Yes/No} & \makecell*[{{L{12.2cm}}}]{Có \textbf{tiền sử} bị bệnh gì \textbf{} không ? (Have a history of any illness?)} & 6.6 \\ 
        \hline
        \makecell[l]{Who} & \makecell*[{{L{12.2cm}}}]{Những \textbf{người nào} dễ bị xơ gan? (Who is susceptible to cirrhosis?)}  & 9.0 \\ 
        \hline
        \makecell[l]{Why} & \makecell*[{{L{12.2cm}}}]{\textbf{Vì sao} nhang có thể ảnh hưởng xấu tới cơ thể? (Why incense can adversely affect the body?)} & 7.8 \\ 
        \hline
        \makecell[l]{Which} & \makecell*[{{L{12.2cm}}}]{\textbf{Nhóm nào} chiếm tỉ lệ cao nhất? (Which group accounts for the highest percentage?)} & 7.0 \\
        \hline
        \makecell[l]{When} & \makecell*[{{L{12.2cm}}}]{\textbf{Khi nào} thì cô có thể kết thúc điều trị? (When can she finish treatment?)}  & 2.6 \\ 
        \hline
        \makecell[l]{Where} & \makecell*[{{L{12.2cm}}}]{Zhou Xiaoying sinh sống \textbf{ở đâu?} (Where does Zhou Xiaoying live?)}  & 4.0 \\ 
        \hline
        \makecell[l]{Others} & \makecell*[{{L{12.2cm}}}]{Còn du thuyền Diamond Princess? Kể tên một số quốc gia có số mắc cao (About the Diamond Princess yacht? Name a few countries with high risk?)} & 5.6 \\
        \hline
    \end{tabular} 
    }
\end{table}

In Vietnamese, the process of interaction contains statements between two people. Each statement contains two functional elements, including the negotiatory for carrying the argument in statements that go through the conversation and the remainder to keep the rest information of statements \cite{thai2004metafunctional}. The negotiatory is an essential part of the statement in the conversation. The negotiatory element comprises interrogatives particles, element interrogatives items, and imperative particles. The interrogatives are the characteristic of questions. In Table \ref{question analysis}, we show all kinds of questions in Vietnamese that are usually used in daily life. The interrogative words are marked bold in the sentence. According to Table \ref{question analysis}, the "What" type accounts for the highest ratio in the UIT-ViCoQA corpus (32.6\%). 

\begin{table}[H]
        \centering
        \caption{Linguistic phenomena in UIT-ViCoQA questions}
        \label{phenomena} 
        \resizebox{\textwidth}{!}{
            \begin{tabular}{L{1.7cm}lR{2.1cm}}
                \hline
               \multicolumn{1}{|L{1.6cm}|}{\textbf{Phenomenon} } & \multicolumn{1}{c|}{\textbf{Example}} & \multicolumn{1}{R{1cm}|}{\textbf{Ratio (\%)}} \\ \hline
                \multicolumn{3}{|c|}{Relationship between a question and its passage} \\ 
                \hline
                \multicolumn{1}{|L{1.6cm}|}{Lexical match} & \multicolumn{1}{l|}{\makecell*[{{L{12cm}}}]{Q: \textbf{Ai} làm giám đốc quốc gia của Hiệp hội Sảy thai? (\textbf{Who} is the director of the association of miscarriage?)\\ A: Ruth Bender - Atik \\ S: Ruth Bender - Atik, giám đốc quốc gia của Hiệp hội Sảy thai (Ruth Bender - Atik, national director of the association of miscarriage)}} & \multicolumn{1}{R{1cm}|}{47.6} \\ \hline
                \multicolumn{1}{|L{1.6cm}|}{Paraphrasing} & \multicolumn{1}{l|}{\makecell*[{{L{12cm}}}]{Q: \textbf{ Giá} cho mỗi con robot là bao nhiêu? (\textbf{How much} is the price of each robot?) \\ A: 500000 RMB \\ S: Các robot có giá 500000 RMB (khoảng 72000 USD) (Robots have price 500000 RBM, about 72000 USD)}} & \multicolumn{1}{R{1cm}|}{48.0} \\ 
                \hline
                \multicolumn{1}{|L{1.6cm}|}{Pragmatics} & \multicolumn{1}{l|}{\makecell*[{{L{12cm}}}]{Q: \textbf{Vì sao?} (\textbf{Why?}) \\ A: Do sầu riêng chứa nhiều chất dinh dưỡng, nhiều năng lượng, cộng với cồn nồng độ cao làm cho nhịp tim tăng (Because durian contains lots of nutrients, energy, combining with high concentration of alcohol, make heartbeat increase.)
                \\ S: Chuyên gia dinh dưỡng Nguyễn Mộc Lan cho biết sầu riêng nhiều chất dinh dưỡng, nhiều năng lượng, cộng với rượu nồng độ cao làm cho nhịp tim tăng.
                (Nutritionist Nguyen Moc Lan said durian has a lot of nutrients, lots of energy, plus a high concentration of alcohol makes your heart rate increase.)}} & \multicolumn{1}{R{1cm}|}{4.4} \\ 
                \hline
                \multicolumn{3}{|c|}{Relationship between a question and its conversation history} \\ 
                \hline
                \multicolumn{1}{|L{1.6cm}|}{No coreference} & \multicolumn{1}{l|}{\makecell*[{{L{12cm}}}]{Q: Phô mai có giá trị dinh dưỡng \textbf{thế nào}? 
                (\textbf{How} does cheese have nutritional value?)}} & \multicolumn{1}{R{1cm}|}{73.6} \\ 
                \hline
                \multicolumn{1}{|L{1.6cm}|}{Explicit coreference} & \multicolumn{1}{l|}{\makecell*[{{L{12cm}}}]{Q1:  Loại bệnh \textbf{nào} Tiểu Lý mắc phải từ ban đầu?
                (\textbf{What} kind of illness was Tieu Ly initially? )
                \\ A1: bệnh lao phổi (tuberculosis) \\ Q2: Anh ta chữa bệnh trong thời gian \textbf{bao lâu?} (\textbf{How long} does he treat?)}} & \multicolumn{1}{R{1cm}|}{20.6} \\ 
                \hline
                \multicolumn{1}{|L{1.6cm}|}{Implicit coreference} & \multicolumn{1}{l|}{\makecell*[{{L{12cm}}}]{Q1: Ở Hải Phòng bệnh nhân \textbf{từ đâu}  trở về? (\textbf{Where} does the patient come from in Hai Phong?) \\ A1: Quảng Đông (Guangdong) \\ Q2: Hiện có triệu chứng \textbf{ gì}?  (\textbf{What} symptoms are there?)}} & \multicolumn{1}{R{1cm}|}{5.8} \\ 
                \hline
            \end{tabular}
        }
\end{table}

Next, we randomly divide our corpus into training, development, and test sets with proportions 70\%, 15\%, and 15\%, respectively. Then, we take 100 articles by random from the development set to analyze and evaluate the corpus, which is called analysis set \cite{reddy-etal-2019-coqa}. We segment texts in the corpus by the Underthesea framework\footnote{\url{ https://github.com/undertheseanlp/underthesea}}. 

According to Gupta et al. \cite{gupta-etal-2020-conversational}, the Conversational Machine Comprehension (CMC) model answers the question by extracting information not only from the reading texts but also from conversational history. Therefore, the main linguistic phenomena in the UIT-ViCoQA are based on the relationship between questions and the reading passage and the relationship between questions and the conversation history. Table \ref{phenomena} displays the linguistic phenomena in the UIT-ViCoQA corpus. 

For the relationship between questions and the reading texts, there are three types of phenomena: lexical match, paraphrasing, and pragmatic. The lexical match indicates that the questions contain the same words as the reading texts. In contrast, paraphrasing is the question in which their words use synonyms from the reading texts, and pragmatic means the question uses words that do not relate to the reading texts. The proportions of lexical match, paraphrasing, and pragmatic phenomenon in the UIT-ViCoQA corpus are 47.6\%, 48.0\%, and 4.4\%, respectively, as shown in Table \ref{phenomena}. 

In addition, for the relationship between questions and the conversation history, there are three types of relational phenomena: no coreference, explicit coreference, and implicit coreference. The percentages of no coreference, explicit coreference, and implicit coreference in the UIT-ViCoQA corpus are 73.6\%, 20.6\%, and 5.8\%, respectively, according to Table \ref{phenomena}.
\section{Methodologies}
\label{method}
According to Gupta et al. \cite{gupta-etal-2020-conversational}, a typical conversation reading comprehension task consists of reading passage as context (C), the conversation history (H) includes multiple question-answer pairs, and the generated answers (A). Therefore, this task combines two models: the machine reading comprehension model for encoding the questions and context into neural space vectors and the question-answering model to generate and decode answers from questions to natural language. 

For the machine reading comprehension model, the Document Reader (DrQA) introduced by Chen et al. \cite{chen-etal-2017-reading} is a powerful model on various of machine reading comprehension corpora such as: SQuAD \cite{rajpurkar-etal-2016-squad}, TextWorldsQA \cite{labutov-etal-2018-multi}, and UIT-ViQuAD \cite{nguyen-etal-2020-vietnamese}. The DrQA model consists of two modules: Document Retriever and Document Reader. We use the Document Reader of the DrQA to extract the answer spans for the questions. 

Besides, for the conversational comprehension task, the generated answers are not only from the reading passage but also the conversation history. The model extracts the history of conversations as a special context to generate new answers. SDNeT model \cite{zhu2019sdnet} is a contextual attention-based model based on the idea of DrQA with a special mechanism to extract the context of the conversation. 

Furthermore, The FLOW mechanism enables the MRC models to encode the history of the conversation comprehensively. Hence, this mechanism integrates well the latent semantic of the conversation history. FlowQA \cite{huang2019flowqa} and GraphFlow \cite{ijcai2020-171} are two flow-based neural models that grasping the conversational history context to generate answers.
\section{Experiments}
\label{experiment}

\subsection{Data preparation}
We pre-process the data before fitting to the model by these following steps: (1) Removing special characters and stop words, (2) Segmenting sentences into words by using the Underthesea tool, and (3) Transforming the texts into vectors by using fastText word embedding in the Vietnamese language provided by Grave et al. \cite{grave-etal-2018-learning}. The dimension of fastText word embedding is 300.

\subsection{Evaluation metrics}
We evaluate the performance of the models by comparing the generated answers with the accurate answers on F1-score and Exact match (EM) score. The F1-score measures the right predicted answers comparing with the correct answers. The EM score measures the exact matching of prediction answers with original answers \cite{rajpurkar-etal-2016-squad}.

\subsection{Experiment results}
The FLOW models give optimistic results on the UIT-ViCoQA corpus. According to Table \ref{tbl_performace}, FlowQA obtains the highest result by F1-score on both development and test sets. 
For the EM score, the SDNet model gives the highest results. However, there is a large gap between the F1 and the EM scores as well as the performance of CMC models and human performance.

\begin{table}[H]
    \centering
    \caption{Experimental results on the UIT-ViCoQA corpus}
    \label{tbl_performace}
    \resizebox{\textwidth}{!}{
    \begin{tabular}{|L{4cm}|R{2cm}|R{2cm}|R{2cm}|R{2cm}|}
        \hline
        \multicolumn{1}{|c|}{\multirow{2}{*}{\textbf{Model}}} & \multicolumn{2}{c|}{\textbf{EM (\%)}}  & \multicolumn{2}{c|}{\textbf{F1-score (\%)}} \\ 
        \cline{2-5} 
        \multicolumn{1}{|c|}{}  & \multicolumn{1}{c|}{\textbf{Dev}} & \multicolumn{1}{c|}{\textbf{Test}} & \multicolumn{1}{c|}{\textbf{Dev}} & \multicolumn{1}{c|}{\textbf{Test}} \\ 
        \hline
        DrQA  & 13.17  & 13.50  & 43.28 & 37.71 \\ 
        \hline
        SDNet & \textbf{15.40} & \textbf{15.60} & 41.90 & 40.50 \\
        \hline
        FlowQA & 13.13 & 12.53 & \textbf{44.84} & \textbf{45.27} \\ 
        \hline
        GraphFlow & 13.77 & 14.73 & 44.69 & 45.16 \\ 
        \hline
        \textbf{Human performance} & \textbf{35.67} & \textbf{38.66} & \textbf{73.33} & \textbf{76.18} \\
        \hline
    \end{tabular}
    }
\end{table}

\subsection{Error analysis}

\begin{table}
    \centering
    \caption{The answers predicted by models on a sample in the UIT-ViCoQA corpus}
    \label{corpus_predicted_analysis} 
    \resizebox{\textwidth}{!}{
        \begin{tabular}{|L{1.5cm}|p{13.5cm}|}
        \hline
        \multicolumn{2}{|p{15cm}|}{Tính đến ngày 18/2, Việt Nam có 16 ca nhiễm covid-19. Trong đó, Vĩnh Phúc có tới 5 công nhân và 6 người thân của họ bị lây nhiễm. Con số này khiến các doanh nghiệp đặt ra câu hỏi về nguy cơ lây lan virus khó lường trong môi trường doanh nghiệp. Chỉ cần một trường hợp phát hiện nhiễm Covid-19 là cả văn phòng, phân xưởng tiếp xúc với người bệnh sẽ phải cách ly cô lập, gây gián đoạn hoạt động sản xuất kinh doanh, tạo áp lực lên hệ thống y tế công. Ông Đoàn Đình Duy Khương - Tổng Giám đốc điều hành Dược Hậu Giang về vấn đề bảo vệ sức khỏe lao động cho biết, mỗi ngày họ phải dành hơn 1/3 thời gian cho nơi làm việc ...
        (Up to 18/2, Vietnam has 16 affected cases of covid-19. Specifically, Vinh Phuc has 5 workers and 6 relatives of whom are affected. This number makes the enterprises question about the risk of virus spreading in working environment. If only one case is detected to be affected Covid-19, the whole offices, factories which are contacted with the patients will be quarantined, disrupting production and business activities, and putting pressure on the public health system. Mr Đoàn Đình Duy Khương - General director of Hau Giang Pharmacy about protecting labor health affairs, says that, everyday they have to spend more than 1/3 of their time at work ... )
        } \\
        \hline
        Q1 & Việt Nam có bao nhiêu ca nhiễm tính đến 18/2? (How many affected cases of Vietnam have been counted up to 2/18?) \\
        Original & Tính đến ngày 18/2, Việt Nam có 16 ca nhiễm covid-19. (Up to 18/2, Vietnam has 16 affected cases of covid-19 )\\
        DrQA & 16 \\
        SDNet & 16 ca nhiễm covid-19. (16 affected cases of covid-19)  \\
        FlowQA & 16 ca (16 cases) \\
        GraphQA & 16 ca nhiễm covid-19 (16 affected cases of covid-19) \\
        \hline
        Q2 & Cụ thể? (Detail?) \\
        Original & Trong đó, Vĩnh Phúc có tới 5 công nhân và 6 người thân của họ bị lây nhiễm. (Specifically, Vinh Phuc has 5 workers and 6 relatives of whom are affected.) \\
        DrQA & Vĩnh Phúc có tới 5 công nhân và 6 người thân của họ bị lây nhiễm. (Vinh Phuc has 5 workers and 6 relatives of whom are affected.) \\
        SDNet & Trong đó, Vĩnh Phúc có tới 5 công nhân và 6 người thân của họ bị lây nhiễm. (Vinh Phuc has 5 workers and 6 relatives of whom are affected.)  \\
        FlowQA & Trong đó, Vĩnh Phúc có tới 5 công nhân và 6 người thân của họ bị lây nhiễm. (Vinh Phuc has 5 workers and 6 relatives of whom are affected.)  \\
        GraphQA & Trong đó, Vĩnh Phúc có tới 5 công nhân và 6 người thân của họ bị lây nhiễm. (Vinh Phuc has 5 workers and 6 relatives of whom are affected.)  \\
        \hline
        Q3 & Doanh nghiệp nghĩ gì? (What does the enterprise think about?)\\
        Original & Con số này khiến các doanh nghiệp đặt ra câu hỏi về nguy cơ lây lan virus khó lường trong môi trường doanh nghiệp.(This number makes the enterprises question about the risk of virus spreading in working environment.)  \\
        DrQA & doanh nghiệp đặt ra câu hỏi về nguy cơ lây lan virus khó lường trong môi trường doanh nghiệp.(the enterprise questions about the risk of virus spreading in working environment.) \\
        SDNet & ca nhiễm covid-19. (affected cases of covid-19)  \\
        FlowQA & các doanh nghiệp đặt ra câu hỏi về nguy cơ lây lan virus khó lường trong môi trường doanh nghiệp.. (the enterprises question about the risk of virus spreading in working environment.)  \\
        GraphQA & virus khó lường trong môi trường doanh nghiệp. (The virus is dangerous in working environment) \\
        \hline
        Q4 & Nguy cơ là gì? (What is risk?) \\
        Original & Chỉ cần một trường hợp phát hiện nhiễm Covid-19 là cả văn phòng, phân xưởng tiếp xúc với người bệnh sẽ phải cách ly cô lập, gây gián đoạn hoạt động sản xuất kinh doanh, tạo áp lực lên hệ thống y tế công. (If only one case is detected to be affected Covid-19, the whole offices, factories which are contacted with the patients will be quarantined, disrupting production and business activities, and putting pressure on the public health system.)  \\
        DrQA & Chỉ cần một trường hợp phát hiện nhiễm Covid-19 là cả văn phòng. (If only one case is detected to be affected Covid-19, the whole offices.) \\
        SDNet & khó phát triển bền vững.  \\
        FlowQA & Chỉ cần một trường hợp phát hiện nhiễm Covid-19 là cả văn phòng. (If only one case is detected to be affected Covid-19, the whole offices.) \\
        GraphQA & Chỉ cần một trường hợp phát hiện nhiễm Covid-19 là cả văn phòng. (If only one case is detected to be affected Covid-19, the whole offices.) \\
        \hline
        Q5 & Đoàn Đình Duy Khương là ai? (Who is Đoàn Đình Duy Khương?)\\
        Original & Ông Đoàn Đình Duy Khương - Tổng Giám đốc điều hành Dược Hậu Giang. (Mr. Đoàn Đình Duy Khương - General director of Hau Giang Pharmacy) \\
        DrQA & Ông Đoàn Đình Duy Khương - Tổng Giám đốc điều hành Dược Hậu Giang về vấn đề bảo vệ sức khỏe lao động cho biết. (Mr Đoàn Đình Duy Khương - General director of Hau Giang Pharmacy about protecting labor health affairs says) \\
        SDNet & khó phát triển bền vững. (hard to develop stably)  \\
        FlowQA & Ông Đoàn Đình Duy Khương - Tổng Giám đốc điều hành Dược Hậu Gia. (Mr Đoàn Đình Duy Khương - General director of Hau Gia) \\
        GraphQA & Ông Đoàn Đình Duy Khương - Tổng Giám đốc điều hành Dược Hậu Giang về vấn đề bảo vệ sức khoẻ. (Mr Đoàn Đình Duy Khương - General director of Hau Giang Pharmacy about protecting health affairs)\\
        \hline
        \end{tabular}
    }
\end{table}

Table \ref{corpus_predicted_analysis} shows the predicted answers given by four different models, including DrQA, SDNet, FlowQA, and GraphFlow, respectively. In general, FlowQA and GraphFlow give the most relevant answer as the original answer. For example, in the question Q3 - "What the enterprise think about?", the reader needs to look back to the previous question-answer Q1-A1 and Q2-A2 to inference the context about the "affected cases of COVID-19" (Q1) and the "detailed of affected cases" (Q2). GraphFlow and FlowQA offer the most relevant answer than DrQA for the question Q3. For question Q5, GraphFlow provides the most relevant answer about the person mentioned in the reading passage, while other models give the answer with redundant information in comparison with the original answer. For the question Q4, both four models cannot give the exact answer. This is due to the ambiguity of Vietnamese interrogative words in questions where it is written in the genuine and non-genuine form. For example, the question Q2: "Cụ thể?" can be understood as "\textbf{What} is the detail?" or "\textbf{How} it happened?". Besides, the question Q4: "Nguy cơ là gì?" can be understood as "\textbf{What} is the risk?" or "\textbf{How} bad is the risk?". This is known as the MOOD in the Vietnamese. The interrogative clause in Vietnamese consists of two main elements: the negotiatory and the remainders. The negotiatory carries the centroid of the interaction. This aspect of Vietnamese interrogative is described carefully by Thai \cite{thai2004metafunctional}.

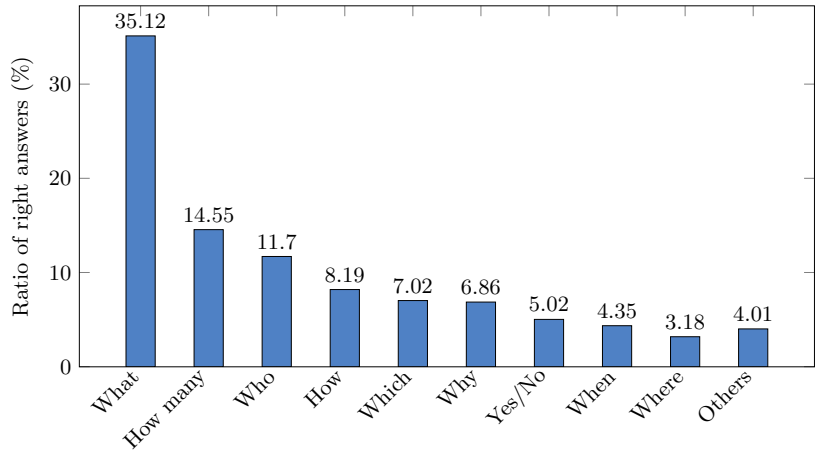
\begin{figure}[H]
    \centering
    \resizebox{.9\textwidth}{!}{
    \begin{tikzpicture}
        \begin{axis}[
            symbolic x coords={What, How many, Who, How, Which, Why, Yes/No, When, Where, Others},
            xtick=data,
            width  = 1*\textwidth,
            height = 6.8cm,
            ylabel = {Ratio of right answers (\%)},
            bar width=12pt,
            nodes near coords,
            nodes near coords align={vertical},
            ylabel near ticks,
        	x tick label
            style={font=\footnotesize,rotate=45, anchor=east}
          ]
            \addplot[ybar,fill=lightblue] coordinates {
                (What,   35.12)
                (How many,  14.55)
                (Who,   11.70)
                (How, 8.19)
                (Which, 7.02)
                (Why, 6.86)
                (Yes/No, 5.02)
                (When, 4.35)
                (Where, 3.18)
                (Others, 4.01)
            };
        \end{axis}
    \end{tikzpicture}
    }
    \caption{The impact of question types on the performance of models.}
    \label{fig_question_type}
\end{figure}

In addition, we study the ability of the models for retrieving correct answers based on the type of questions on the development set. Figure \ref{fig_question_type} shows the ratio of correct answers by different kinds of questions in the UIT-ViCoQA corpus. A question gives the right answers if the F1-score is greater than 70\%. According to Figure \ref{fig_question_type}, the question type "What" has the highest ratio, which is 35.12\%. Besides, the question type "What" accounts for 32.6\% as described in Table \ref{question analysis}. Therefore, the models mostly give the correct answers to this kind of question. Furthermore, the question types "How many" and "Who" also have a high ratio.  

\begin{table}[H]
    \centering
    \caption{Types of predicted answer given by the models}
    \label{tbl_answer_analyze} 
    \resizebox{\textwidth}{!}{
        \begin{tabular}{|L{1.5cm}|C{4cm}|C{7.6cm}|R{1cm}|}
        \hline
        \textbf{Types} & \textbf{Description} & \textbf{Example} & \textbf{Ratio (\%)} \\
        \hline
        Matching answers & \makecell*[{{L{4cm}}}]{The predicted answers fully match with truth answers} & \makecell*[{{L{7.5cm}}}]{Q: Việc này có giúp tình trạng tốt lên không? (Does this help improve the condition?) \\ P: Không (No) \\ A: Không (No)} & 16.73 \\
        \hline
        Free-form answers & \makecell*[{{L{4cm}}}]{The predicted answer only match the a part of truth answers}  &  \makecell*[{{L{7.5cm}}}]{Q:Tỷ lệ ung thu Việt Nam có cao không? (Is the rate of cancer in Vietnam high?)  \\ P: cao (High)  \\ A: có (Yes) } & 59.93 \\
        \hline
        Wrong answers & \makecell*[{{L{4cm}}}]{The predicted answer does not match the truth answer} & \makecell*[{{L{7.5cm}}}]{Q: Béo phì có gây dậy thì sớm không? (Does obesity cause early puberty?)  \\ P: Không (No)  \\ A: Có (Yes) } & 23.27 \\
        \hline
    \end{tabular}
    }
\end{table}

Finally, we analyze the predicted answers on the development set. According to Table \ref{tbl_answer_analyze}, there are three types of the answer given by the models, and most of the predicted answers are concentrated on the free-form type, which accounts for 59.93\%. This is why the F1 and EM scores have a considerable difference, as described in Table \ref{tbl_performace}.

In general, most error predictions are due to the number of questions and the variety of answers, as well as the linguistic phenomena. Therefore, it is necessary to increase the number of questions and the question types as well as enriching answers to make the corpus more diverse. 
\section{Conclusion and future work}
In this paper, we propose the dataset about machine reading comprehension for healthcare texts in Vietnamese. This dataset includes 2,000 health articles with 10,000 questions. We also conduct experiments on several baseline models, and the best result in the F1-score is 45.27\%. Nevertheless, the difference between F1 and EM scores is large. This is due to the linguistic phenomena about the Vietnamese interrogative particles and the limited answers. Therefore, it is necessary to increase the number of questions and answers as well as make questions and answers more diverse in further research. Besides, enabling the CMC models to capture and understand the contextual meaning of the conversation history is also a challenging task in the conversational machine reading comprehension model researching. 

In future, we plan to increase the quantity and quality of the UIT-ViCoQA corpus as well as to conduct further experiments on deep learning and transfer learning using pre-trained language models \cite{devlin2018bert,rogers2020primer,conneau2019unsupervised,nguyen2020phobert} to enhance the performance of CMC models on the UIT-ViCoQA corpus. Inspired by the conversational question answering system \cite{qu2020open}, we suggest using this model and UIT-ViCoQA for building Vietnamese conversational question answering systems.

\label{conclusion}

\subsubsection*{Acknowledgements}
We would like to express our thanks to reviewers for their valuable comments to help improve our work. Besides, we would like to thank our annotators for their cooperation.  

\bibliographystyle{splncs04}
\bibliography{reference}

\end{document}